\documentclass[letterpaper]{article}
\usepackage{aaai}
\usepackage{times}
\usepackage{helvet}
\usepackage{courier}
\usepackage{amsmath,pgfmath,pgffor}
\DeclareMathOperator{\erf}{erf}
\DeclareMathOperator{\swish}{swish}
\usepackage{graphicx}
\usepackage{booktabs}
\usepackage[table]{xcolor}
\usepackage{color, colortbl}
\definecolor{LightCyan}{rgb}{0.9,0.95,0.97}
\newcolumntype{a}{>{\columncolor{LightCyan}}c}
\begin{document}

\title{SERF: Towards better training of deep neural networks using log-Softplus ERror activation Function}

% \author{Anonymous Authors*}

\author{Sayan Nag*,\textsuperscript{1}
Mayukh Bhattacharyya*,\textsuperscript{2}\\\\
\textsuperscript{1}{University of Toronto}\\
\textsuperscript{2}{Stony Brook University}\\
}

\maketitle
\begin{abstract}
\begin{quote}
Activation functions play a pivotal role in determining the training dynamics and neural network performance. The widely adopted activation function ReLU despite being simple and effective has few disadvantages including the \textit{Dying ReLU} problem. In order to tackle such problems, we propose a novel activation function called Serf which is self-regularized and non-monotonic in nature. Like Mish, Serf also belongs to the Swish family of functions. Based on several experiments on computer vision (image classification and object detection) and natural language processing (machine translation, sentiment classification and multi-modal entailment) tasks with different state-of-the-art architectures, it is observed that Serf vastly outperforms ReLU (baseline) and other activation functions including both Swish and Mish, with a markedly bigger margin on deeper architectures. Ablation studies further demonstrate that Serf based architectures perform better than those of Swish and Mish in varying scenarios, validating the effectiveness and compatibility of Serf with varying depth, complexity, optimizers, learning rates, batch sizes, initializers and dropout rates. Finally, we investigate the mathematical relation between Swish and Serf, thereby showing the impact of pre-conditioner function ingrained in the first derivative of Serf which provides a regularization effect making gradients smoother and optimization faster.

\end{quote}
\end{abstract}

\section{Introduction}

Activation functions are point-wise functions which play a crucial role in introducing non-linearity in neural networks. In a neural network, linear transformed inputs are passed through an activation function giving rise to non-linear counterparts. These non-linear point-wise activation functions are hugely responsible for the performance of neural networks. Thus, choosing a suitable activation function for better training and improved efficiency has always been an interesting area of research. Activation functions like $tanh$ and $sigmoid$ were widely used in previous works \shortcite{lecun1989backpropagation,lecun1998gradient,mira1995natural,jarrett2009best}. However, they had some disadvantages including upper-boundedness. This paved the way for the development of an activation function widely known as Rectified Linear Unit (ReLU) \shortcite{nair2010rectified}. Being simple yet effective, ReLU was not only easier to optimize compared to its contemporaries (sigmoid and tanh), but also showed better generalization and improved convergence properties, which led to its wide adoption.

ReLU however has few disadvantages including the infamous \textit{Dying ReLU} phenomenon \shortcite{lu2019dying,maas2013rectifier}. The absence of any negative portion resulted in such a problem which can be noticed through a gradient information loss caused by collapsing the negative inputs to zero. On the other hand, ReLU is also non-differentiable which can result in inconsistencies during gradient-based optimization. Keeping those in mind, researchers have propounded various activation functions including leaky ReLU \shortcite{maas2013rectifier}, PReLU\shortcite{he2015delving}, ELU \shortcite{clevert2015fast}, GELU \shortcite{hendrycks2016gaussian}, SELU \shortcite{klambauer2017self}, Swish \shortcite{ramachandran2017searching}, Mish \shortcite{misra2019mish}, etc. Out of the aforementioned activation functions, Mish mostly outperforms its contemporaries including Swish. Mish has a continuous profile which renders better information propapagation as compared to ReLU. It was inspired by the self-gating property of Swish. As opposed to Swish, Mish possess a preconditioner which results in smoother gradients and better optimization.

In this work, we have proposed a novel activation function called \textit{Serf} which is non-monotonic and is also inspired by the self-gating property of Swish. We define Serf as $f(x) = x \erf(\ln(1+e^{x}))$ where $erf$ is the error function \shortcite{andrews1998special}. Swish, Mish and Serf belong to the same family of activation functions possessing self-gating property. Like Mish, Serf also possess a pre-conditioner which results in better optimization and thus enhanced performance. Our experiments demonstrate that our proposed activation function Serf outperforms ReLU, Swish and even Mish for different standard architectures in a variety of tasks including image classification, object detection, graph node classification, machine translation, sentiment classification and multi-modal entailment, involving varied datasets. We have also conducted ablation studies on MNIST \shortcite{lecun1998mnist} and CIFAR-10 datasets \cite{cifar10} to demonstrate the efficiency of Serf over Swish and Mish.

\section{Related Work}

One of the mostly used activation functions is Rectified Linear Units (ReLU) \shortcite{nair2010rectified}. Originally proposed for Restricted Boltzmann Machines, this activation function gained prominence because of its simplicity and effectiveness and eventually replaced the sigmoid and tanh units. Despite being computationally efficient, it is not entirely devoid of shortcomings. In order to address those, Leaky ReLU (LReLU) was introduced which replaced the constant zero portion of the ReLU function with a linear function thereby 'leaking' some information \shortcite{maas2013rectifier}. LReLU showed superior performance compared to ReLU, and the performance was further enhanced when the slope of the negative part was learnt as an extra parameter using Parametric ReLU (PReLU) \shortcite{he2015delving}. However, lower boundedness is important in order to render strong regularization effects which was absent in both LReLU and PReLU. Furthermore, similar to ReLU, they are also not differentiable.

Keeping these aspects in mind, researchers proposed activation functions like Exponetial Linear Units (ELU) \shortcite{clevert2015fast} and Scaled Exponential Linear Units (SELU) \shortcite{klambauer2017self}. ELU and SELU possess better convergence characteristics along with a saturation plateau in its negative region. However, these activation functions have found to be incompatible with Batch Normalization (BN) \shortcite{ioffe2015batch}.

Finally, using self-gating property Swish was proposed which addressed the aforementioned drawbacks to a greater extent abreast demonstrating superior results compared to the previous established activation functions \shortcite{ramachandran2017searching}. Belonging to the same class as Swish, another activation function called Mish was proposed which performed equally well or better than Swish in most of the computer vision tasks \shortcite{misra2019mish}. Our proposed activation function, Serf, is also inspired from the self-gating mechanism and thus belongs to the Swish-like class of functions. It has been shown experimentally that our proposed Serf outperforms other activation functions in a variety of computer vision and natural language processing tasks.

\section{Serf}

\subsection{Motivation}

Activation functions introduce non-linearity in the neural networks and they play a very important role in the overall performance of a network. ReLU has been the most widely used activation function in neural networks. However, it suffers from several disadvantages, the most noticeable one being the \textit{dying ReLU} phenomenon. This problem ensued from the missing negative part in the ReLU activation function which restrains the negative values to zero. At the same time, ReLU is \textit{not} continuously differentiable. Furthermore, ReLU is a non-negative function. This creates a \textit{non-zero mean} problem where the mean activation larger than zero. Such an issue is not desirable for network convergence \shortcite{clevert2015fast}.

\begin{figure*}[t!]
    \centering
    \includegraphics[width = \textwidth, height = 0.27\textwidth]{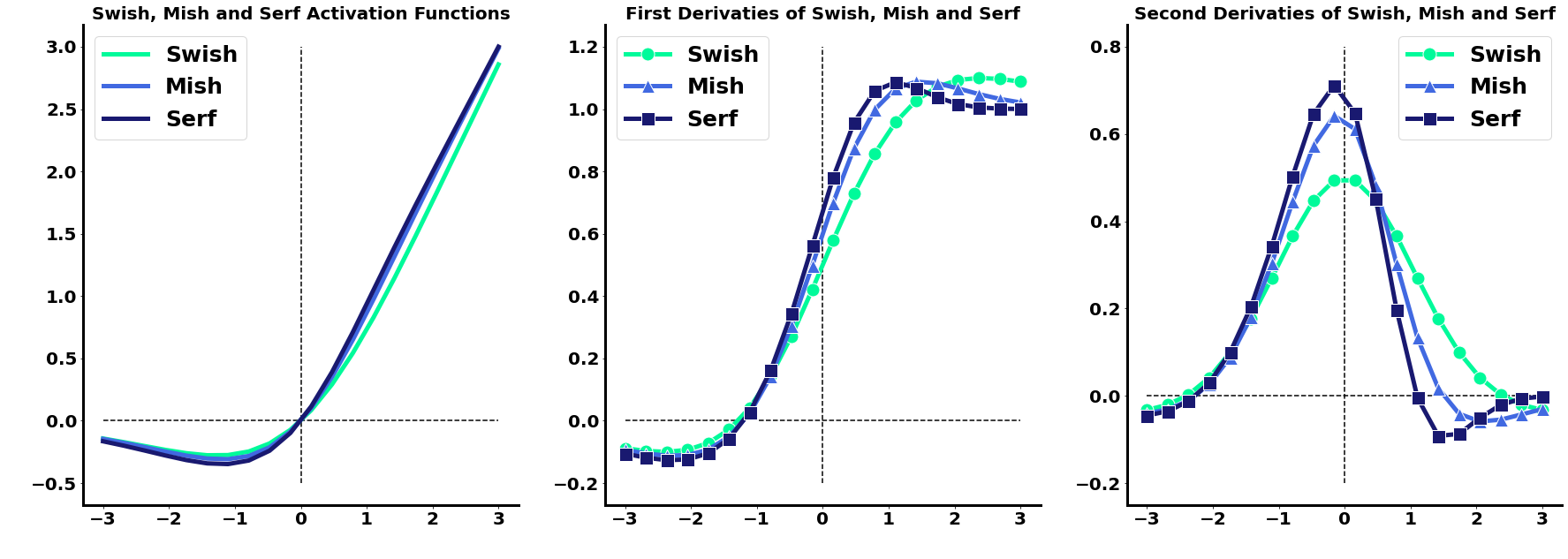}
    \caption{Activation functions (Left), first derivatives (Middle) and second derivatives (Right) for Swish, Mish and Serf.}%
    \label{fig:Activation Functions}
\end{figure*}

In order to address these aforesaid problems to some extent, in the recent past several new activation functions emerged including leaky ReLU, ELU, Swish, etc. Swish is seemingly an ideal candidate for an activation function with properties including non-monotonicity and ability to preserve small negative weights abreast maintaining a smooth profile. Similar to Swish, activation functions like GELU \shortcite{hendrycks2016gaussian} has gained popularity especially in the transformer based architectures used both in the fields of Computer Vision (ViT \shortcite{dosovitskiy2020image} and MLP Mixer \shortcite{tolstikhin2021mlp}), as well as Natural Language Processing (GPT-2 \shortcite{radford2019language} and GPT-3 \shortcite{brown2020language}). Another activation function which rose to prominence due to its performance in state-of-the-art classification and object detection tasks, is Mish. Mish has its roots in Swish and was developed by methodical analysis over the attributes that led to the efficacy of Swish.

Taking inspiration from the development of Mish, we propose an activation function called Serf. Serf is defined as:

\begin{equation}
    f(x) = x \erf(\ln(1+e^{x}))
\end{equation}

\subsection{Properties}

Serf is bounded below and unbounded above. Serf is smooth, non-monotonic and differentiable. It also preserves small portion of negative weights. Serf is inspired by Swish and Mish where the \textit{self-gating} property has been used to multiply the output of a non-linear function of an input with the same non-modulated input. Self-gating is advantageous because it requires only a single-scalar input, whilst normal gating requires multiple two-scalar inputs \shortcite{ramachandran2017searching}.

\begin{itemize}

\item \textbf{Upper unboundedness}: Activation functions like $tanh$ and $sigmoid$ have upper bounds. So, initialization should happen in the linear regime of these activation functions. Such a property is not desirable since it leads to saturation while training  due to near-zero gradients \shortcite{glorot2010understanding}. ReLU being unbounded above attempted to avoid the saturation problem. This is a crucial attribute which can be noticed in all the successors of the ReLU function like leakyReLU, GELU, Swish, Mish, etc. Serf also possesses this feature, with its positive side as an approximate linear function of the input (see Figure \ref{fig:Activation Functions}). This makes Serf a good candidate for an activation function.

\item \textbf{Lower boundedness}: Activation functions must posses lower bounds in order to provide strong regularization effects. However, in the ReLU activation function a neuron receiving negative input will always output zero eventually become \textit{dead} or inactive and hence useless. This is referred to as the \textit{dying ReLU} phenomenon \shortcite{lu2019dying,maas2013rectifier}. It usually happens when the learning rate is high or if there is a large negative bias. By preserving of a small portion of negative information, Serf mitigates the aforementioned problem furthermore resulting in better expressivity and improved gradient flow. The negative bound for Serf is approximately 0.3484 (see Figure \ref{fig:Activation Functions}).

\item \textbf{Differentiability}: Unlike ReLU, Serf is continuously differentiable. This is beneficial owing to the fact that it avoids singularities and any concomitant ill-effects during gradient-based optimization.

\item \textbf{Preconditioner}: Serf is closely related to Swish which can be noticed in its first derivative. The first derivative of Serf is given as:

\begin{equation}
    \begin{split}
        f'(x) = \frac{2}{\sqrt{\pi}}e^{-\ln((1+e^{x}))^{2}} x \sigma(x) + \frac{f(x)}{x}\\
        = p(x)\swish(x) + \frac{f(x)}{x}
    \end{split}
\end{equation}

Here, $\sigma$ is the sigmoid function and $p(x)$ is a preconditioner function. Preconditoners make the gradients smoother and have been previously used extensively in optimization problems. The inverse of a symmetric positive definite matrix has been used as a preconditioner in case of gradient descent. Application of such preconditioners makes the objective function smoother thereby increasing the rate of convergence \shortcite{axelsson1986rate}. Therefore, the strong regularization effect contributed by such preconditoner in case of Serf makes the gradients smoother and optimization faster, thereby outperforming Swish as can be noticed in the experiments.

Mish also has a precondtioner which makes it perform better than Swish. The difference between Mish and Swish is that in Serf we used the error function ($\erf$) whereas in Mish $tanh$ function is used. Serf, however, outperforms Mish in most experiments (see Experiments and Ablations). We speculate that Serf's preconditioner function renders better regularization effects than that of Mish.

\begin{figure}[t!]
    \centering
    \includegraphics[width = 0.23\textwidth,height = 0.23\textwidth]{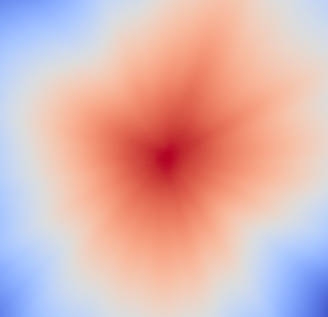}
    \hspace{0cm}
    %\includegraphics[width = 0.25\textwidth,height = 0.25\textwidth]{mish_mnist_2.png}
    %\hspace{1cm}
    \includegraphics[width = 0.23\textwidth,height = 0.23\textwidth]{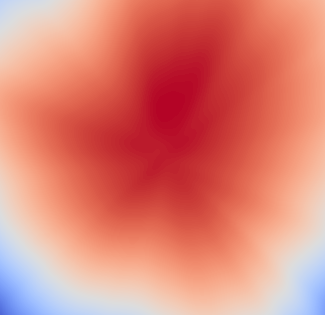}
    \caption{ Output landscapes of a randomly initialized 6-layered neural network with ReLU (Left) and Serf (Right) activations.}%
    \label{fig:loss landscapes}
\end{figure}

\item \textbf{Smoothness}: Smooth loss landscapes indicates easier optimization with less local optima and hence better generalization minimizing influence of initializations and learning rates. The output landscapes of a randomly chosen 6-layered neural network with ReLU and Serf activation functions has been shown in Figure \ref{fig:loss landscapes}. It is to be noted that output landscape is indicative of the loss landscape. We randomly initialize a 6-layered neural network, where we pass the x and y coordinates of each point in a grid as input, and plot the scalar network output for each grid point. For ReLU activation function, the output landscape of the neural network has sharp transitions in contrast to that of Serf. This conforms to the enhanced performance of Serf as compared to ReLU.

\end{itemize}

\begin{figure*}[t!]
    \centering
    \includegraphics[width = \textwidth]{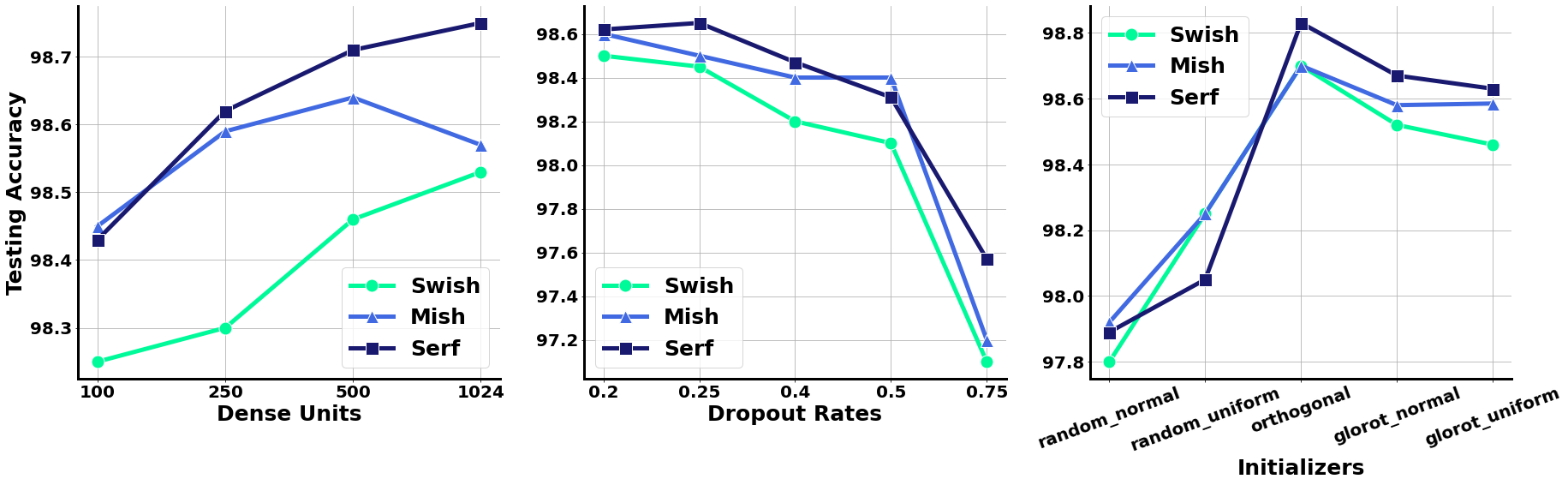}
    \includegraphics[width = \textwidth, height = 0.27\textwidth]{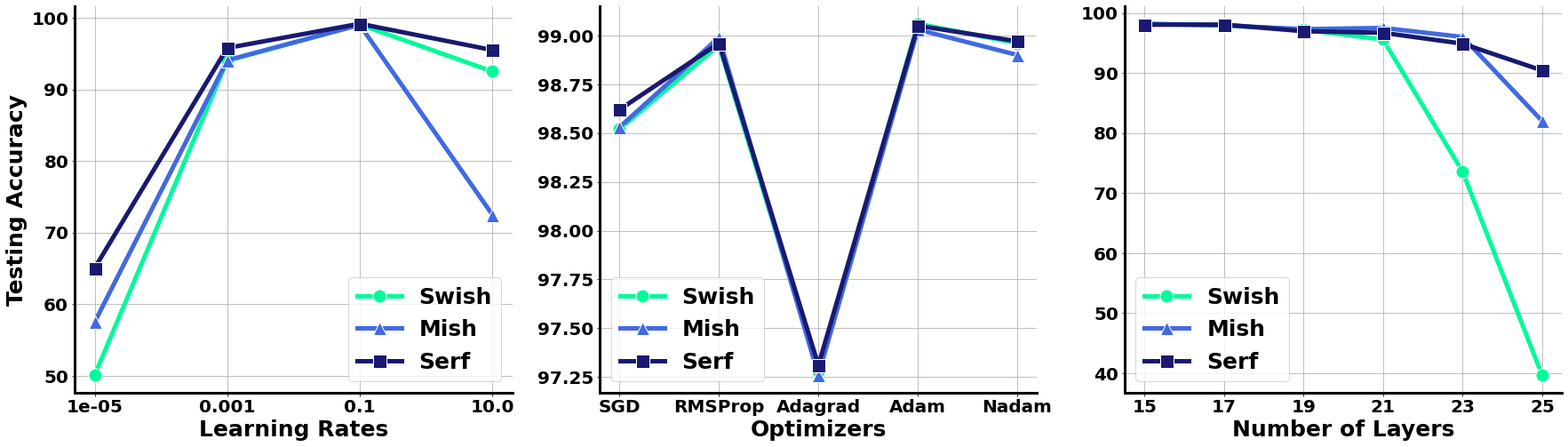}
    \caption{Ablations for MNIST dataset. Top: Testing Accuracies vs Dense Units (Left), Dropout Rates (Middle) and Initializers (Right) for Swish, Mish and Serf. Bottom: Testing Accuracies vs Learning Rates (Left), Optimizers (Middle) and Number of Layers (Right) for Swish, Mish and Serf.}%
    \label{fig:MNIST Ablations}
\end{figure*}

\section{Experiments}

In this section we will demonstrate the performance of our proposed activation function, Serf when used in different state-of-the-art architectures on image, sequence and graph datasets for disparate tasks. We will also present results from ablation studies done on MNIST and CIFAR-10 datasets. Overall, our proposed activation function, Serf outperformed its contemporaries in most of the tasks.

\subsection{Ablations}

Model hyper-parameters play an important role in the training and optimization processes of a neural network thus having direct consequences in the generalizability of a network. Such hyper-parameters include network depths, network widths, type of weight initializations, dropout rates, batch sizes, learning rates, and optimizers. Here we analyze and compare the impacts of different hyper-parameters on our chosen networks with three different activation functions namely Swish, Mish, and Serf. We have used MNIST and CIFAR-10 datasets for this purpose.

\subsubsection{MNIST}

\begin{itemize}

\item \textbf{Dense Units}: The number of dense units refers to the number of neurons present in a dense layer. In this case we have used a 4 layered architecture with one dense layer followed by a Batch Normalization Layer and SGD \shortcite{robbins1951stochastic} as an optimizer. We observe that as the number of dense units increases, the model complexity increases and Serf outperforms Swish and Mish (Fig \ref{fig:MNIST Ablations}). This suggests that Serf works well with complex models. This has also been noticed in other experiments.

\item \textbf{Dropout Rates}: As the dropout rate increases, the overall performance for all three activation functions drop, however, the performance degradation for Serf is relatively lesser as compared to Swish and Mish (Fig \ref{fig:MNIST Ablations}).

\item \textbf{Initializers}: The performance of Serf is better than both Swish and Mish in all except for random uniform initialization (Fig \ref{fig:MNIST Ablations}). This suggests that Serf is a better candidate compared to its contemporaries.

\item \textbf{Learning Rates}: With varying learning rates, Serf performs better than both Swish and Mish (Fig \ref{fig:MNIST Ablations}). Particularly, with higher learning rates, the degradation is quite pronounced in Swish and not that much in Mish and Serf. We have used SGD \shortcite{robbins1951stochastic} as an optimizer in this case.

\item \textbf{Optimizers}: In this case, with varying optimizers, the overall performance of Serf is equal or marginally better than Swish and Mish (Fig \ref{fig:MNIST Ablations}). Performance drop can be noticed for all three activation functions in case of Adagrad optimizer \shortcite{duchi2011adaptive}.

\item \textbf{Number of layers}: In this case, each dense layer was followed by a Batch Normalization layer. As the number of dense layers increases, i.e., the depth of the network increases, models become complex and optimization becomes difficult. The degradation in the performances for all the three different activation functions conforms the aforementioned fact. However, Serf maintained a significantly higher accuracy as compared to Swish and Mish (Fig \ref{fig:MNIST Ablations}). This makes Serf a suitable candidate for large and complex networks.

% \footnote{Code: https://anonymous.4open.science/r/Serf-3630/}

\end{itemize}

\begin{figure*}[t!]
    \centering
    \includegraphics[width = \textwidth, height = 0.27\textwidth]{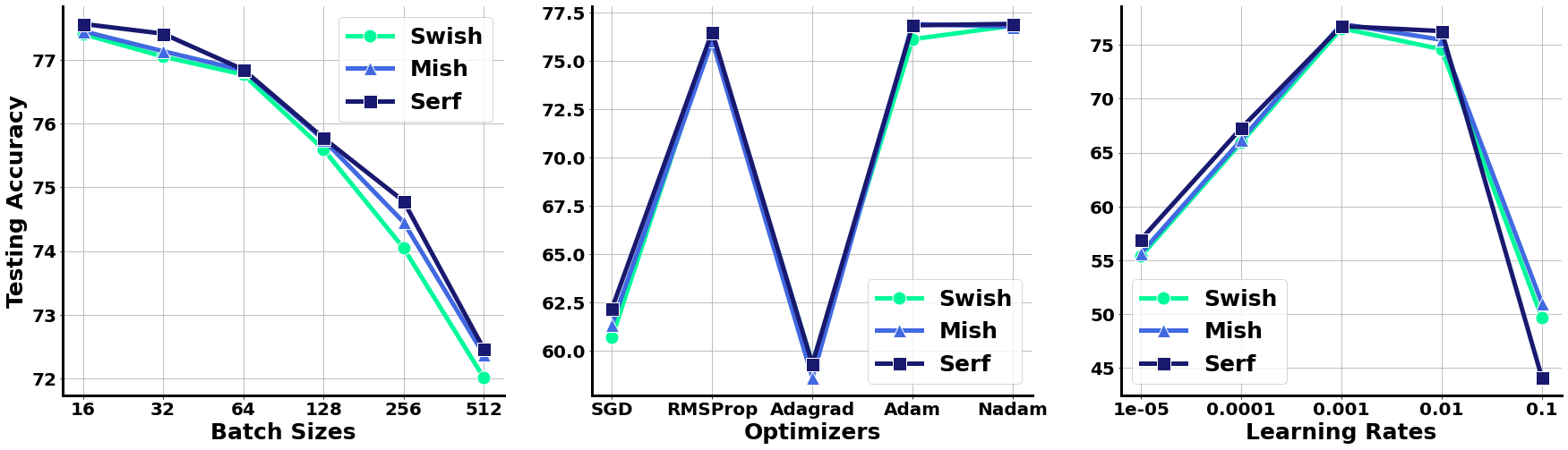}
    \caption{Ablations for CIFAR-10 dataset. Testing Accuracies vs Batch Sizes (Left), Optimizers (Middle) and Learning Rates (Right) for Swish, Mish and Serf.}%
    \label{fig:CIFAR-10 Ablations}
\end{figure*}

\subsubsection{CIFAR-10}
We have used a ResNet-18 model with a dense layer and a classification head in tandem. The results are obtained with training the model without pre-trained weights over multiple runs for 20 epochs, which gives a decent convergence point.
\begin{itemize}

\item \textbf{Batch Size}: We observed that with decreasing training batch sizes, the performance for all the competing activation functions drop (Fig \ref{fig:CIFAR-10 Ablations}), however, Serf holds the better position out of all three over all the batch sizes. Adam \shortcite{kingma2014adam} is used as the optimzer here.

\item \textbf{Optimizers}: In this case, with varying optimizers, the overall performance of Serf is equal or marginally better than Swish and Mish (Fig \ref{fig:CIFAR-10 Ablations}). Performance drop can be noticed for all three activation functions in case of  \shortcite{duchi2011adaptive} and SGD optimizers \shortcite{robbins1951stochastic}.

\item \textbf{Learning Rates}: Serf is observed to be performing better or equal to both Mish and Swish on all of the learning rates evaluated barring 0.1 where a steeper drop is observed in Serf compared to the other two activation functions (Fig \ref{fig:CIFAR-10 Ablations}). Adam \shortcite{kingma2014adam} is used as the optimzer here.

\end{itemize}

\subsection{Image Classification}

For image classification we have considered different standard architectures applied on two datasets namely CIFAR-10 and CIFAR-100.

\subsubsection{CIFAR-10/100:}

%We have considered a batch size of 128, Adam optimizer with a learning rate of 0.01 (without any scheduler) and have trained each architecture for 50 epochs.

We have considered different deep learning architectures (for CIFAR-10: SqueezeNet \shortcite{iandola2016squeezenet}, Resnet-50 \shortcite{he2016deep}, WideResnet-50-2 \shortcite{zagoruyko2016wide}, ShuffleNet-v2 \shortcite{ma2018shufflenet}, ResNeXt-50 \shortcite{xie2017aggregated}, Inception-v3 \shortcite{szegedy2016rethinking}, DenseNet-121 \shortcite{huang2017densely}, MobileNet-v2 \shortcite{sandler2018mobilenetv2} and EfficientNet-B0 \shortcite{tan2019efficientnet}; for CIFAR-100: Resnet-164 \shortcite{he2016identity}, WideResnet-28-10 \shortcite{zagoruyko2016wide}, DenseNet-40-12 \shortcite{huang2017densely}, Inception-v3 \shortcite{szegedy2016rethinking}) with three disparate activation functions, namely, ReLU (baseline), Mish and Serf (proposed). This has been done for image classification task on CIFAR-10 and CIFAR-100 datasets where for each network we have only changed the activation functions and have kept every other parameter constant for fair comparisons. Tables \ref{tab:sota_cifar10} and \ref{tab:sota_cifar100} show that Serf consistently outperformed both ReLU and Mish activation functions across all the architectures used in the experiment (1\% to 2\% improvement over baseline ReLU) for both CIFAR-10 and CIFAR-100 datasets, thereby suggesting Serf to be the best candidate activation function for CIFAR-10/100 image classification tasks.

\begin{table}[htp]\footnotesize
    \centering
    \renewcommand{\arraystretch}{1.3}
    \begin{tabular}{ccca}

    \toprule
    \textbf{Methods} & \textbf{ReLU} & \textbf{Mish} & \textbf{Serf (Ours)}\\[.5ex]
    \midrule
    
    SqueezeNet & 84.14 & 85.98 & \textbf{86.32}\\[.5ex]
    
    Resnet-50 & 86.54 & 87.03 & \textbf{88.07}\\[.5ex]
    
    WideResnet-50-2 & 86.39 & 86.57 & \textbf{86.73}\\[.5ex]
    
    ShuffleNet-v2 & 83.93 & 84.07 & \textbf{84.55}\\[.5ex]
    
    ResNeXt-50 (32 $\times$ 4d) & 87.25 & 87.97 & \textbf{88.49}\\[.5ex]
    
    Inception-v3 & 90.93 & 91.55 & \textbf{92.89}\\[.5ex]
    
    DenseNet-121 &  88.59 & 89.05 & \textbf{89.07}\\[.5ex]
     
    MobileNet-v2 & 85.74 & 86.39 & \textbf{86.61}\\[.5ex]
    
    EfficientNet-B0 (Swish) & 78.26 & 78.02 & \textbf{78.41}\\[.5ex]
    
    \bottomrule
    \end{tabular}\\
    \caption{Top-1 \% Accuracy values of different state-of-the-art methods for different activation functions on CIFAR-10}
    \label{tab:sota_cifar10}
    
    % \bigskip
    
    % \begin{tabular}{c|c c}

    % \hline
    % Activations & Top-1 \% Accuracy & Top-5 \% Accuracy\\[.5ex]
    % \hline
    % \hline
    
    % MLP-Mixer (GELU) &  64.14 & \textbf{96.71}\\[.5ex]
    
    % MLP-Mixer (Serf) & \textbf{64.36} & 96.59\\[.5ex]
    
    % \hline
    
    % CCT (ReLU) &  79.05 &  98.65\\[.5ex]
    
    % CCT (Mish) &  80.02 &  98.70\\[.5ex]
    
    % CCT(Serf) & \textbf{80.23} & \textbf{97.72}\\[.5ex]
    
    % \hline
    % \end{tabular}\\[0.5ex]
    % \caption{Top-1 and Top-5 \% Accuracy values (classification) after 10 epochs of training MLP-Mixer for GELU (SOTA) and Serf activation functions on CIFAR-10 test dataset and 50 epochs of training Compact Convolutional Transformer (CCT) for ReLU, Mish and Serf functions on CIFAR-10 test dataset.}
    % \label{tab:MLP_mixer_CCT}
    
\end{table}

\begin{table}[ht]\footnotesize
    \centering
    \renewcommand{\arraystretch}{1.3}
    \begin{tabular}{ccca}

    \toprule
    \textbf{Methods} & \textbf{ReLU} & \textbf{Mish} & \textbf{Serf (Ours)}\\[.5ex]
    \midrule
    
    Resnet-164 & 74.55 & 75.02 & \textbf{75.13}\\[.5ex]
    
    WideResnet-28-10 & 76.32 & 77.03 & \textbf{77.54}\\[.5ex]
    
    DenseNet-40-12 & 73.68 & 73.91 & \textbf{74.16}\\[.5ex]
    
    Inception-v3 & 71.54 & 72.38 & \textbf{72.95}\\[.5ex]
    
    \bottomrule
    \end{tabular}\\[0.5ex]
    \caption{Top-1 \% Accuracy values of different state-of-the-art methods for different activation functions on CIFAR-100}
    \label{tab:sota_cifar100}
\end{table}

\begin{table}[ht]\footnotesize
    \centering
    \renewcommand{\arraystretch}{1.3}

    \begin{tabular}{ccc}

    \toprule
    \textbf{Activations} & \textbf{Top-1 \% Acc} & \textbf{Top-5 \% Acc}\\[.5ex]
    \midrule
    
    MLP-Mixer (GELU) &  64.14 & \textbf{96.71}\\[.5ex]
    
    \rowcolor{LightCyan}
    
    MLP-Mixer (Serf) & \textbf{64.36} & 96.59\\[.5ex]
    
    \midrule
    
    CCT (ReLU) &  79.05 &  97.72\\[.5ex]
    
    CCT (Mish) &  80.02 &  \textbf{98.70}\\[.5ex]
    
    \rowcolor{LightCyan}
    
    CCT(Serf) & \textbf{80.23} & 98.65\\[.5ex]
    
    \bottomrule
    \end{tabular}\\
    \caption{Top-1 and Top-5 \% Accuracy values (classification) after 10 epochs of training MLP-Mixer for GELU (SOTA) and Serf activation functions on CIFAR-10 test dataset and 50 epochs of training Compact Convolutional Transformer (CCT) for ReLU, Mish and Serf functions on CIFAR-10 test dataset.}
    \label{tab:MLP_mixer_CCT}
    
\end{table}

% \begin{table}[htp]\footnotesize
%     \centering
%     \renewcommand{\arraystretch}{1.3}
%     \begin{tabular}{c|c}

%     \hline
%     Activations & Top-1 \% Accuracy\\[.5ex]
%     \hline
%     \hline
    
%     ReLU & 42.22\\[.5ex]
    
%     Mish & 45.13\\[.5ex]
    
%     Serf & \textbf{45.98}\\[.5ex]
    
%     \hline
%     \end{tabular}\\[0.5ex]
%     \caption{Top-1 \% Accuracy values of a simple involution for different activation functions on CIFAR-10}
%     \label{tab:involution_simple}
% \end{table}

We have also used two recent architectures namely MLP Mixer \shortcite{tolstikhin2021mlp} and Compact Convolutional Transformers (CCT) \shortcite{hassani2021escaping}. We have trained the MLP Mixer for 10 epochs only on CIFAR-10 training dataset, separately with two different activation functions (GELU and Serf) and tested the models on the CIFAR-10 test dataset. The Top-1 \% and Top-5 \% Accuracy values (see Table \ref{tab:MLP_mixer_CCT}) suggest that Serf's performance is comparable to GELU's (baseline) performance for MLP Mixer. We have also trained CCT for 50 epochs on CIFAR-10 training dataset, separately with three different activation functions (ReLU, Mish and Serf) and tested the models on CIFAR-10 test dataset. Serf clearly outperforms ReLU (baseline) and Mish in this case (see Table \ref{tab:MLP_mixer_CCT}). The results indicate that Serf is a better activation function for transformer based architectures.

% \begin{table}[htp]\footnotesize
%     \centering
%     \renewcommand{\arraystretch}{1.3}
%     \begin{tabular}{c|c}

%     \hline
%     Activations & Top-1 \% Accuracy\\[.5ex]
%     \hline
%     \hline
    
%     GELU &  \textbf{58.99}\\[.5ex]
    
%     Serf & 58.93\\[.5ex]
    
%     \hline
%     \end{tabular}\\[0.5ex]
%     \caption{Top-1 \% Accuracy values (classification) of ViT for GeLU (SOTA) and Serf activation functions on CIFAR-10 test dataset.}
%     \label{tab:vit}
% \end{table}

% \subsubsection{ImageNet-1k:}

% *Need to fill up*

\subsection{Object Detection}

Object detection is considered one of the important visual scene understanding tasks. In our case, we have considered Pascal VOC dataset for object detection task using YOLOv3 \shortcite{redmon2018yolov3} and tiny YOLOv3 frameworks. Leaky ReLU is intrinsic to the YOLOv3 framework. For fair comparisons we have only changed the activation function keeping other hyperparameters fixed as outlined in \shortcite{redmon2018yolov3}.
%In each case, we have considered 150 epochs since that leads to a good enough convergence. 
Mean Average Precision (MAP) scores in Table \ref{tab:obj_detection} clearly indicate that our proposed Serf outperforms the baseline leaky ReLU based architectures in the object detection task for Pascal VOC dataset.

\begin{table}[htp]\footnotesize
    \centering
    \renewcommand{\arraystretch}{1.3}
    \begin{tabular}{cccc}

    \toprule
    \textbf{Model} & \textbf{Activations} & \textbf{MAP@.5} & \textbf{MAP@.5:.95}\\[.5ex]
    \midrule
    
    YOLOv3 & LeakyReLU & 0.740 & 0.473\\[.5ex]
    
    \rowcolor{LightCyan}
    
    YOLOv3 & Serf (Ours) & \textbf{0.766} & \textbf{0.501}\\[.5ex]
     
    \midrule
    
    YOLOv3 Tiny & LeakyReLU & 0.503 & 0.219\\[.5ex]
    
    \rowcolor{LightCyan}
    
    YOLOv3 Tiny & Serf (Ours) & \textbf{0.514} & \textbf{0.227}\\[.5ex]

    \bottomrule
    \end{tabular}\\
    \caption{Mean Average Precision scores for different object detection models on the Pascal VOC dataset. LeakyReLU is instrinsic to YOLO framework.}
    \label{tab:obj_detection}
\end{table}

\subsection{Semi-supervised Node Classification}

Following the implementations outline in \shortcite{kipf2016semi}, we have considered 3 different datasets, namely CITESEER, CORA and PUBMED for semi-supervised node classification using three disparate activation functions, namely, ReLU (baseline), Mish and Serf (proposed). All training parameters and hyper-parameters have been kept same as mentioned in \shortcite {kipf2016semi} for fair comparisons. Table. \ref{tab:sota_gnn} shows that Serf either performed equally well or better than both ReLU and Mish activation functions across the three different datasets, thereby indicating versatility of the proposed activation function.

\begin{table}[htp]\footnotesize
    \centering
    \renewcommand{\arraystretch}{1.3}
    \begin{tabular}{ccca}

    \toprule
    \textbf{Dataset} & \textbf{ReLU} & \textbf{Mish} & \textbf{Serf (Ours)}\\[.5ex]
    \midrule
    
    CORA & 81.5 & \textbf{81.7} & \textbf{81.7}\\[.5ex]
    
    CITESEER & 70.3 & 71.3 & \textbf{71.7}\\[.5ex]
    
    PUBMED & 79.0 & 79.3 & \textbf{79.4}\\[.5ex]
    
    \bottomrule
    \end{tabular}\\[0.5ex]
    \caption{Top-1 \% Accuracy values of different state-of-the-art GNN semi-supervised node classification methods for different activation functions on CORA, CITESEER and PUBMED.}
    \label{tab:sota_gnn}
\end{table}

\subsection{Machine Translation, Sentiment Classification \& Multi-modal Entailment}

In this section, we have demonstrated the effectiveness of our proposed Serf activation function in Machine Translation and Sentiment Classification tasks. We have considered 3 different architectures and datasets.

For Machine Translation we have used a sequence-to-sequence Transformer \shortcite{vaswani2017attention} Encoder-Decoder based model trained (for 20 epochs) on the the Multi30k dataset for German-English translation \shortcite{W16-3210}. For comparison purposes, we have considered ReLU, GELU, Mish and Serf (proposed) and observed that Serf outperformed the remaining three activation functions as suggested by the BLEU scores shown in the Table \ref{tab:transformer_seq2seq}.

For sentiment classification, we have considered two datasets, namely imdb movie review sentiment and Pol Emo 2.0 sentiment datasets. For the imdb movie review sentiment dataset \shortcite{maas-EtAl:2011:ACL-HLT2011}, we have considered: (i) a simple architecture consisting of a 1D conv net with a text embedding layer which we have trained using three different activation functions and noticed that Serf outperformed the other two activation functions (ReLU and Mish) suggesting that Serf also works well with simple architectures (see Table \ref{tab:nlp_simple}), and (ii) a 4-layer Transformer model which we have also trained for 20 epochs for each of the three activation functions eventually obtaining the best results for the proposed Serf function (see Table \ref{tab:nlp_simple}). For the Pol Emo 2.0 sentiment database \shortcite{kocon2019polemo}, we have trained a BERT based model \shortcite{devlin2018bert} for two different activation functions, Mish and Serf. The Precision, Recall and F1-scores suggest that Serf performed equally well or better than Mish for this task (see Table \ref{tab:polBERT}).

For multi-modal entailment task, we have used the multi-modal entailment database, recently introduced by the Google Research \footnote{https://github.com/google-research-datasets/recognizing-multimodal-entailment}. We have used a smaller variant of the original BERT model. The code used for this purpose is available at \footnote{https://github.com/sayakpaul/Multimodal-Entailment-Baseline}. We have used two activation functions for comparison purposes: GELU and Serf. Table \ref{tab:multimodal_entailment} shows the accuracies on test dataset averaged over 5 runs (each trained for 10 epochs). The accuracy values suggest that Serf performs marginally better than GELU in this case.

\begin{table}[ht!]\footnotesize
    \centering
    \renewcommand{\arraystretch}{1.3}
    \begin{tabular}{c c c c a}

    \toprule
    \textbf{Scores} & \textbf{ReLU} & \textbf{GELU} & \textbf{Mish} & \textbf{Serf (Ours)}\\[.5ex]
    \midrule
    
    BLEU & 35.55 & 35.62 & 35.36 & \textbf{36.06}\\[.5ex]
    
    \bottomrule
    \end{tabular}\\[0.5ex]
    \caption{BLEU scores of seq2seq Transformer model (after training for 20 epochs) for different activation functions on Multi30k test dataset.}
    \label{tab:transformer_seq2seq}
    
    \bigskip
    
    \begin{tabular}{c c c a}

    \toprule
    \textbf{Model} & \textbf{ReLU} & \textbf{Mish} & \textbf{Serf (Ours)}\\[.5ex]
    \midrule
    
    1D conv with text-embedding & 85.36 & 85.99 & \textbf{86.18}\\[.5ex]
    
    4-layer Transformer model & 88.82 & 88.99 & \textbf{89.03}\\[.5ex]
    
    \bottomrule
    \end{tabular}\\
    \caption{Top-1 \% Accuracy values of 1D conv net with a text embedding layer and 4-layer Transformer model for ReLU, Mish and Serf on imdb movie review sentiment dataset.}
    
    \label{tab:nlp_simple}
    
    \bigskip
    
    \begin{tabular}{cccc}

    \toprule
    \textbf{Activation} & \textbf{Precision} & \textbf{Recall} & \textbf{F1-score}\\[.5ex]
    \midrule
    
    Mish & 0.8374 & 0.8329 & \textbf{0.8346}\\[.5ex]
    
    \rowcolor{LightCyan}
    
    Serf (Ours) & \textbf{0.8377} &  \textbf{0.8330} & 0.8342\\[.5ex]
    
    \bottomrule
    \end{tabular}\\
    \caption{Precision, Recall and F1-scores for different activation functions for sentiment classification using BERT on Pol Emo 2.0 sentiment database.}
    
    \label{tab:polBERT}
    
    \bigskip
    
    \begin{tabular}{c c a}

    \toprule
    \textbf{Metric} & \textbf{GELU} & \textbf{Serf (Ours)}\\[.5ex]
    \midrule
    
    Mean Accuracy & 85.28 & \textbf{85.42}\\[.5ex]
    
    \bottomrule
    \end{tabular}\\
    \caption{Mean Accuracy values of GELU and Serf based architectures for multi-modal entailment task.}
    
    \label{tab:multimodal_entailment}
    
\end{table}

%\subsection{Stability, Accuracy and Trade-Off}

\section{Conclusions and Future Works}

In this paper, we have proposed a novel activation function called Serf. Serf has properties such as upper unboundedness, lower boundedness, non-monotonicity and smoothness which are the desired properties for an activation function. Serf has shown to be effective against different ablations compared to its contemporaries like Swish and Mish. Experimental results with different state-of-the-art architectures on varied datasets for disparate tasks including image classification, object detection, sentiment classification, machine translation and multi-modal entailment demonstrate that the proposed Serf outperforms the baseline ReLU performance and as well as other activation functions like Swish, Mish and GELU. The results can be improved with desired hyperparameters for Serf which can be obtained with a hyperparameter search.

Future works include: (1) the understanding of the importance and contribution of pre-conditioner as a regularizer and how modifying it can have an impact on the final results; this can lead to the development of more effective activation functions, (2) the development of \textit{Hard-Serf} like Hard-Swish and Hard-Mish and compare its performance on different benchmark datasets and tasks, (3) the development and exploration of probabilistic version of Serf as shown in \shortcite{lee2019probact}, (4) the development of parameterized Serf like PReLU, and finally (5) comparison the performance of Serf with other contemporary activation functions for tasks such as image super-resolution, image reconstruction, etc. Overall, Serf is a simple, effective and versatile activation function which can be incorporated in any neural network for better training and performance gains.

\bibliographystyle{aaai}
\fontsize{9.0pt}{10.0pt}
\bibliography{ref}

\begin{thebibliography}{}

\bibitem[\protect\citeauthoryear{Andrews}{1998}]{andrews1998special}
Andrews, L.~C.
\newblock 1998.
\newblock {\em Special functions of mathematics for engineers}, volume~49.
\newblock Spie Press.

\bibitem[\protect\citeauthoryear{Axelsson and
  Lindskog}{1986}]{axelsson1986rate}
Axelsson, O., and Lindskog, G.
\newblock 1986.
\newblock On the rate of convergence of the preconditioned conjugate gradient
  method.
\newblock {\em Numerische Mathematik} 48(5):499--523.

\bibitem[\protect\citeauthoryear{Brown \bgroup et al\mbox.\egroup
  }{2020}]{brown2020language}
Brown, T.~B.; Mann, B.; Ryder, N.; Subbiah, M.; Kaplan, J.; Dhariwal, P.;
  Neelakantan, A.; Shyam, P.; Sastry, G.; Askell, A.; et~al.
\newblock 2020.
\newblock Language models are few-shot learners.
\newblock {\em arXiv preprint arXiv:2005.14165}.

\bibitem[\protect\citeauthoryear{Clevert, Unterthiner, and
  Hochreiter}{2015}]{clevert2015fast}
Clevert, D.-A.; Unterthiner, T.; and Hochreiter, S.
\newblock 2015.
\newblock Fast and accurate deep network learning by exponential linear units
  (elus).
\newblock {\em arXiv preprint arXiv:1511.07289}.

\bibitem[\protect\citeauthoryear{Devlin \bgroup et al\mbox.\egroup
  }{2018}]{devlin2018bert}
Devlin, J.; Chang, M.-W.; Lee, K.; and Toutanova, K.
\newblock 2018.
\newblock Bert: Pre-training of deep bidirectional transformers for language
  understanding.
\newblock {\em arXiv preprint arXiv:1810.04805}.

\bibitem[\protect\citeauthoryear{Dosovitskiy \bgroup et al\mbox.\egroup
  }{2020}]{dosovitskiy2020image}
Dosovitskiy, A.; Beyer, L.; Kolesnikov, A.; Weissenborn, D.; Zhai, X.;
  Unterthiner, T.; Dehghani, M.; Minderer, M.; Heigold, G.; Gelly, S.; et~al.
\newblock 2020.
\newblock An image is worth 16x16 words: Transformers for image recognition at
  scale.
\newblock {\em arXiv preprint arXiv:2010.11929}.

\bibitem[\protect\citeauthoryear{Duchi, Hazan, and
  Singer}{2011}]{duchi2011adaptive}
Duchi, J.; Hazan, E.; and Singer, Y.
\newblock 2011.
\newblock Adaptive subgradient methods for online learning and stochastic
  optimization.
\newblock {\em Journal of machine learning research} 12(7).

\bibitem[\protect\citeauthoryear{Elliott \bgroup et al\mbox.\egroup
  }{2016}]{W16-3210}
Elliott, D.; Frank, S.; Sima'an, K.; and Specia, L.
\newblock 2016.
\newblock Multi30k: Multilingual english-german image descriptions.
\newblock In {\em Proceedings of the 5th Workshop on Vision and Language},
  70--74.
\newblock Association for Computational Linguistics.

\bibitem[\protect\citeauthoryear{Glorot and
  Bengio}{2010}]{glorot2010understanding}
Glorot, X., and Bengio, Y.
\newblock 2010.
\newblock Understanding the difficulty of training deep feedforward neural
  networks.
\newblock In {\em Proceedings of the thirteenth international conference on
  artificial intelligence and statistics},  249--256.
\newblock JMLR Workshop and Conference Proceedings.

\bibitem[\protect\citeauthoryear{Hassani \bgroup et al\mbox.\egroup
  }{2021}]{hassani2021escaping}
Hassani, A.; Walton, S.; Shah, N.; Abuduweili, A.; Li, J.; and Shi, H.
\newblock 2021.
\newblock Escaping the big data paradigm with compact transformers.
\newblock {\em arXiv preprint arXiv:2104.05704}.

\bibitem[\protect\citeauthoryear{He \bgroup et al\mbox.\egroup
  }{2015}]{he2015delving}
He, K.; Zhang, X.; Ren, S.; and Sun, J.
\newblock 2015.
\newblock Delving deep into rectifiers: Surpassing human-level performance on
  imagenet classification.
\newblock In {\em Proceedings of the IEEE international conference on computer
  vision},  1026--1034.

\bibitem[\protect\citeauthoryear{He \bgroup et al\mbox.\egroup
  }{2016a}]{he2016deep}
He, K.; Zhang, X.; Ren, S.; and Sun, J.
\newblock 2016a.
\newblock Deep residual learning for image recognition.
\newblock In {\em Proceedings of the IEEE conference on computer vision and
  pattern recognition},  770--778.

\bibitem[\protect\citeauthoryear{He \bgroup et al\mbox.\egroup
  }{2016b}]{he2016identity}
He, K.; Zhang, X.; Ren, S.; and Sun, J.
\newblock 2016b.
\newblock Identity mappings in deep residual networks.
\newblock In {\em European conference on computer vision},  630--645.
\newblock Springer.

\bibitem[\protect\citeauthoryear{Hendrycks and
  Gimpel}{2016}]{hendrycks2016gaussian}
Hendrycks, D., and Gimpel, K.
\newblock 2016.
\newblock Gaussian error linear units (gelus).
\newblock {\em arXiv preprint arXiv:1606.08415}.

\bibitem[\protect\citeauthoryear{Huang \bgroup et al\mbox.\egroup
  }{2017}]{huang2017densely}
Huang, G.; Liu, Z.; Van Der~Maaten, L.; and Weinberger, K.~Q.
\newblock 2017.
\newblock Densely connected convolutional networks.
\newblock In {\em Proceedings of the IEEE conference on computer vision and
  pattern recognition},  4700--4708.

\bibitem[\protect\citeauthoryear{Iandola \bgroup et al\mbox.\egroup
  }{2016}]{iandola2016squeezenet}
Iandola, F.~N.; Han, S.; Moskewicz, M.~W.; Ashraf, K.; Dally, W.~J.; and
  Keutzer, K.
\newblock 2016.
\newblock Squeezenet: Alexnet-level accuracy with 50x fewer parameters and< 0.5
  mb model size.
\newblock {\em arXiv preprint arXiv:1602.07360}.

\bibitem[\protect\citeauthoryear{Ioffe and Szegedy}{2015}]{ioffe2015batch}
Ioffe, S., and Szegedy, C.
\newblock 2015.
\newblock Batch normalization: Accelerating deep network training by reducing
  internal covariate shift.
\newblock In {\em International conference on machine learning},  448--456.
\newblock PMLR.

\bibitem[\protect\citeauthoryear{Jarrett \bgroup et al\mbox.\egroup
  }{2009}]{jarrett2009best}
Jarrett, K.; Kavukcuoglu, K.; Ranzato, M.; and LeCun, Y.
\newblock 2009.
\newblock What is the best multi-stage architecture for object recognition?
\newblock In {\em 2009 IEEE 12th international conference on computer vision},
  2146--2153.
\newblock IEEE.

\bibitem[\protect\citeauthoryear{Kingma and Ba}{2014}]{kingma2014adam}
Kingma, D.~P., and Ba, J.
\newblock 2014.
\newblock Adam: A method for stochastic optimization.
\newblock {\em arXiv preprint arXiv:1412.6980}.

\bibitem[\protect\citeauthoryear{Kipf and Welling}{2016}]{kipf2016semi}
Kipf, T.~N., and Welling, M.
\newblock 2016.
\newblock Semi-supervised classification with graph convolutional networks.
\newblock {\em arXiv preprint arXiv:1609.02907}.

\bibitem[\protect\citeauthoryear{Klambauer \bgroup et al\mbox.\egroup
  }{2017}]{klambauer2017self}
Klambauer, G.; Unterthiner, T.; Mayr, A.; and Hochreiter, S.
\newblock 2017.
\newblock Self-normalizing neural networks.
\newblock In {\em Proceedings of the 31st international conference on neural
  information processing systems},  972--981.

\bibitem[\protect\citeauthoryear{Koco{\'n}, Za{\'s}ko-Zieli{\'n}ska, and
  Mi{\l}kowski}{2019}]{kocon2019polemo}
Koco{\'n}, J.; Za{\'s}ko-Zieli{\'n}ska, M.; and Mi{\l}kowski, P.
\newblock 2019.
\newblock Polemo 2.0 sentiment analysis dataset for conll.

\bibitem[\protect\citeauthoryear{Krizhevsky, Nair, and Hinton}{}]{cifar10}
Krizhevsky, A.; Nair, V.; and Hinton, G.
\newblock Cifar-10 (canadian institute for advanced research).

\bibitem[\protect\citeauthoryear{LeCun \bgroup et al\mbox.\egroup
  }{1989}]{lecun1989backpropagation}
LeCun, Y.; Boser, B.; Denker, J.~S.; Henderson, D.; Howard, R.~E.; Hubbard, W.;
  and Jackel, L.~D.
\newblock 1989.
\newblock Backpropagation applied to handwritten zip code recognition.
\newblock {\em Neural computation} 1(4):541--551.

\bibitem[\protect\citeauthoryear{LeCun \bgroup et al\mbox.\egroup
  }{1998}]{lecun1998gradient}
LeCun, Y.; Bottou, L.; Bengio, Y.; and Haffner, P.
\newblock 1998.
\newblock Gradient-based learning applied to document recognition.
\newblock {\em Proceedings of the IEEE} 86(11):2278--2324.

\bibitem[\protect\citeauthoryear{LeCun}{1998}]{lecun1998mnist}
LeCun, Y.
\newblock 1998.
\newblock The mnist database of handwritten digits.
\newblock {\em http://yann. lecun. com/exdb/mnist/}.

\bibitem[\protect\citeauthoryear{Lee \bgroup et al\mbox.\egroup
  }{2019}]{lee2019probact}
Lee, J.; Shridhar, K.; Hayashi, H.; Iwana, B.~K.; Kang, S.; and Uchida, S.
\newblock 2019.
\newblock Probact: A probabilistic activation function for deep neural
  networks.
\newblock {\em arXiv preprint arXiv:1905.10761} 5:13.

\bibitem[\protect\citeauthoryear{Lu \bgroup et al\mbox.\egroup
  }{2019}]{lu2019dying}
Lu, L.; Shin, Y.; Su, Y.; and Karniadakis, G.~E.
\newblock 2019.
\newblock Dying relu and initialization: Theory and numerical examples.
\newblock {\em arXiv preprint arXiv:1903.06733}.

\bibitem[\protect\citeauthoryear{Ma \bgroup et al\mbox.\egroup
  }{2018}]{ma2018shufflenet}
Ma, N.; Zhang, X.; Zheng, H.-T.; and Sun, J.
\newblock 2018.
\newblock Shufflenet v2: Practical guidelines for efficient cnn architecture
  design.
\newblock In {\em Proceedings of the European conference on computer vision
  (ECCV)},  116--131.

\bibitem[\protect\citeauthoryear{Maas \bgroup et al\mbox.\egroup
  }{2011}]{maas-EtAl:2011:ACL-HLT2011}
Maas, A.~L.; Daly, R.~E.; Pham, P.~T.; Huang, D.; Ng, A.~Y.; and Potts, C.
\newblock 2011.
\newblock Learning word vectors for sentiment analysis.
\newblock In {\em Proceedings of the 49th Annual Meeting of the Association for
  Computational Linguistics: Human Language Technologies},  142--150.
\newblock Portland, Oregon, USA: Association for Computational Linguistics.

\bibitem[\protect\citeauthoryear{Maas \bgroup et al\mbox.\egroup
  }{2013}]{maas2013rectifier}
Maas, A.~L.; Hannun, A.~Y.; Ng, A.~Y.; et~al.
\newblock 2013.
\newblock Rectifier nonlinearities improve neural network acoustic models.
\newblock In {\em Proc. icml}, volume~30, ~3.
\newblock Citeseer.

\bibitem[\protect\citeauthoryear{Mira and Sandoval}{1995}]{mira1995natural}
Mira, J., and Sandoval, F.
\newblock 1995.
\newblock {\em From Natural to Artificial Neural Computation: International
  Workshop on Artificial Neural Networks, Malaga-Torremolinos, Spain, June 7-9,
  1995: Proceedings}, volume 930.
\newblock Springer Science \& Business Media.

\bibitem[\protect\citeauthoryear{Misra}{2019}]{misra2019mish}
Misra, D.
\newblock 2019.
\newblock Mish: A self regularized non-monotonic neural activation function.
\newblock {\em arXiv preprint arXiv:1908.08681} 4:2.

\bibitem[\protect\citeauthoryear{Nair and Hinton}{2010}]{nair2010rectified}
Nair, V., and Hinton, G.~E.
\newblock 2010.
\newblock Rectified linear units improve restricted boltzmann machines.
\newblock In {\em Icml}.

\bibitem[\protect\citeauthoryear{Radford \bgroup et al\mbox.\egroup
  }{2019}]{radford2019language}
Radford, A.; Wu, J.; Child, R.; Luan, D.; Amodei, D.; Sutskever, I.; et~al.
\newblock 2019.
\newblock Language models are unsupervised multitask learners.
\newblock {\em OpenAI blog} 1(8):9.

\bibitem[\protect\citeauthoryear{Ramachandran, Zoph, and
  Le}{2017}]{ramachandran2017searching}
Ramachandran, P.; Zoph, B.; and Le, Q.~V.
\newblock 2017.
\newblock Searching for activation functions.
\newblock {\em arXiv preprint arXiv:1710.05941}.

\bibitem[\protect\citeauthoryear{Redmon and Farhadi}{2018}]{redmon2018yolov3}
Redmon, J., and Farhadi, A.
\newblock 2018.
\newblock Yolov3: An incremental improvement.
\newblock {\em arXiv preprint arXiv:1804.02767}.

\bibitem[\protect\citeauthoryear{Robbins and
  Monro}{1951}]{robbins1951stochastic}
Robbins, H., and Monro, S.
\newblock 1951.
\newblock A stochastic approximation method.
\newblock {\em The annals of mathematical statistics}  400--407.

\bibitem[\protect\citeauthoryear{Sandler \bgroup et al\mbox.\egroup
  }{2018}]{sandler2018mobilenetv2}
Sandler, M.; Howard, A.; Zhu, M.; Zhmoginov, A.; and Chen, L.-C.
\newblock 2018.
\newblock Mobilenetv2: Inverted residuals and linear bottlenecks.
\newblock In {\em Proceedings of the IEEE conference on computer vision and
  pattern recognition},  4510--4520.

\bibitem[\protect\citeauthoryear{Szegedy \bgroup et al\mbox.\egroup
  }{2016}]{szegedy2016rethinking}
Szegedy, C.; Vanhoucke, V.; Ioffe, S.; Shlens, J.; and Wojna, Z.
\newblock 2016.
\newblock Rethinking the inception architecture for computer vision.
\newblock In {\em Proceedings of the IEEE conference on computer vision and
  pattern recognition},  2818--2826.

\bibitem[\protect\citeauthoryear{Tan and Le}{2019}]{tan2019efficientnet}
Tan, M., and Le, Q.
\newblock 2019.
\newblock Efficientnet: Rethinking model scaling for convolutional neural
  networks.
\newblock In {\em International Conference on Machine Learning},  6105--6114.
\newblock PMLR.

\bibitem[\protect\citeauthoryear{Tolstikhin \bgroup et al\mbox.\egroup
  }{2021}]{tolstikhin2021mlp}
Tolstikhin, I.; Houlsby, N.; Kolesnikov, A.; Beyer, L.; Zhai, X.; Unterthiner,
  T.; Yung, J.; Keysers, D.; Uszkoreit, J.; Lucic, M.; et~al.
\newblock 2021.
\newblock Mlp-mixer: An all-mlp architecture for vision.
\newblock {\em arXiv preprint arXiv:2105.01601}.

\bibitem[\protect\citeauthoryear{Vaswani \bgroup et al\mbox.\egroup
  }{2017}]{vaswani2017attention}
Vaswani, A.; Shazeer, N.; Parmar, N.; Uszkoreit, J.; Jones, L.; Gomez, A.~N.;
  Kaiser, {\L}.; and Polosukhin, I.
\newblock 2017.
\newblock Attention is all you need.
\newblock In {\em Advances in neural information processing systems},
  5998--6008.

\bibitem[\protect\citeauthoryear{Xie \bgroup et al\mbox.\egroup
  }{2017}]{xie2017aggregated}
Xie, S.; Girshick, R.; Doll{\'a}r, P.; Tu, Z.; and He, K.
\newblock 2017.
\newblock Aggregated residual transformations for deep neural networks.
\newblock In {\em Proceedings of the IEEE conference on computer vision and
  pattern recognition},  1492--1500.

\bibitem[\protect\citeauthoryear{Zagoruyko and
  Komodakis}{2016}]{zagoruyko2016wide}
Zagoruyko, S., and Komodakis, N.
\newblock 2016.
\newblock Wide residual networks.
\newblock {\em arXiv preprint arXiv:1605.07146}.

\end{thebibliography}

\end{document}